# Une ontologie pour les systèmes multi-agents ambiants dans les villes intelligentes

Nathan Aky* **, Denis Payet*,
Sylvain Giroux**, Rémy Courdier*

* Laboratoire de Mathématiques et d'Informatique, Université de la Réunion,
Parc Technologique Universitaire - Bâtiment 2 - 2 rue Joseph Wetzell
97490 Sainte-Clotilde - France
{nathan.aky, denis.payet, remy.courdier}@univ-reunion.fr
** Laboratoire DOMUS, Département d'Informatique,
Université de Sherbrooke
2500 boul. Université - Sherbrooke, QC - Canda, J1K 2R1
{Nathan.Aky, Sylvain.Giroux}@usherbrooke.ca

**Abstract.** Actuellement, les villes se dotent de très nombreux équipements connectés dans l'optique de se transformer en "villes intelligentes". Pour piloter cette masse d'objets connectés, on peut y adosser des entités logicielles autonomes, appelées agents, qui vont coopérer et utiliser ces appareils afin d'offrir des services personnalisés. Cependant, cette infrastructure d'objet nécessite d'être structuré sémantiquement pour être exploitée. C'est pourquoi la proposition de cet article est une ontologie, formatée en OWL, décrivant les infrastructures d'objets, leurs liens avec l'organisation du système multi-agents ainsi que les services à délivrer en fonction des utilisateurs du système. L'ontologie est appliquée à la mobilité intelligente pour les personnes à mobilité réduite, et pourra être adaptée aux autres axes de la ville intelligente.

## 1. Introduction

L'évolution vers les villes intelligentes fait partie de la stratégie "Europe2020" (Parlement européen, 2014). C'est une tendance qui va conduire à l'installation hétérogène d'appareils connectés dans le but de mettre en place des services unitaires et clivés. Ceux-ci sont rarement interconnectés, ce qui bride les possibilités offertes par ces installations.

Une manière de répondre à ce problème d'hétérogénéité et de flexibilité consiste en l'utilisation d'entités logicielles autonomes, appelées « agents ». N'étant pas intrinsèquement et exclusivement lié à un appareil, un agent peut en exploiter et en combiner plusieurs afin d'offrir des services plus complexes (S.D. Ramchurn *et al*., 2004) et (J. Soldatod *et al*., 2007).

Cependant, il devient nécessaire de décrire cet ensemble d'appareils de manière structurée et sémantique, dans l'objectif de pouvoir y appliquer des algorithmes de raisonnement (M. Hadzic *et al*. 2009). Puisque ces agents sont autonomes et distribués sur les nœuds d'un réseau, cela implique un phénomène de décentralisation. En outre, il est nécessaire que ces connaissances structurées soient suffisamment légères pour être intégrées directement au sein des agents. Dans ce cadre, notre proposition porte sur une ontologie explicitant et structurant la relation entre les services proposés par les agents et les objets nécessaires à leurs fonctionnements. Pour offrir des services personnalisés, l'ontologie structure également la liaison entre les services à offrir et entre le contexte et les préférences des utilisateurs.

En effet, cette ontologie se destine à être intégrée dans la plateforme multi-agents que nous développons. Cette dernière s'inscrit dans un projet ayant pour objectif d'amener les déplacements des habitants de la ville vers une mobilité intelligente. Plus précisément, il s'agit d'un outil de simulation hybride permettant d'accompagner autant les décideurs chargés de mettre en place des aménagements publics, que directement l'accompagnement de personnes à mobilité réduite, en les mettant directement en relation avec un réseau d'entraide.

La prochaine section vise à présenter les travaux proches, notamment des ontologies, traitants de l'intelligence ambiante, des objets connectés, de l'assistance ou encore des villes intelligentes et des systèmes multi-agents. La section 3, portera sur l'ontologie proposée, la méthodologie employée pour sa construction, les concepts principaux abordés ainsi que son originalité, ce qui constitue le cœur de la proposition. Ensuite, nous proposerons un cas d'exemple et comment l'intégrer au sein de l'ontologie. Finalement, cet article s'achèvera par les perspectives et une synthèse.

## 2. Travaux connexes

Le travail préliminaire à la construction de notre ontologie a consisté en l'étude de travaux et d'ontologies liés aux thématiques abordées, à savoir l'in-telligence ambiante, l'assistance à la personne, l'internet des objets ou les systèmes multi-agents.

Aucune ontologie existante ne couvre l'ensemble de ces thématiques, tout en répondant aux contraintes techniques de notre projet, que sont la décentralisation et la faible capacité de calculs des appareils déployés avec la plateforme.

Cependant, nous avons identifié plusieurs travaux connexes lors de l'étude préliminaire à la construction de notre ontologie. Les sous-sections suivantes présentent ces travaux par thématiques.

## 2.1. Ontologies pour l'intelligence ambiante et l'internet des objets

De nombreux travaux portants sur l'intelligence ambiante et l'utilisation de l'internet des objets ont démontré l'intérêt de la modélisation sémantique des concepts de ces domaines.

Le concept de ''contexte'' est une notion clé qu'il est possible de retrouver dans l'article (X.H. Wang *et al.* 2004) qui fait référence sur cette question. Les auteurs y présentent «CONON», une ontologie dont le concept principal est le contexte. Ils présentent le contexte comme étant composé d'entités informatiques (services, applications, appareils, etc.), d'une localisation (extérieur ou intérieur), de personnes et d'activités.

Dans une autre forme, es travaux de (A. Gomez-Goiri *et al.* 2014) se basent sur une ontologie pour former un espace d'échange de données virtuel, et les règles de communications qui en découlent. Cet espace agit à la façon d'un middleware où plusieurs applications d'intelligence ambiante peuvent coopérer.

Dans la même lignée, les auteurs de (A.K.A. Babu, and R. Sivakumar. 2014.) s'appuient sur une ontologie permettant la description et le raisonnement sur un contexte. Ces connaissances sont utilisées au sein d'un middleware sensible au contexte, pour le fonctionnement d'applications d'intelligence ambiante.

Une approche similaire est présentée dans l'article de (D. Ntalassha *et al.* 2016), qui propose une ontologie, décrite en OWL, pour modéliser le contexte dans une solution basée sur l'Internet des Objets. Pour cela, les auteurs décrivent un «IoTContext» comme étant composé d'un ensemble de contexte : un contexte d'utilisateur, d'un contexte d'appareils, un contexte de système, un contexte temporel et environnemental.

Il est intéressant de noter que l'ensemble de ces études représentent des similitudes sur la représentation et/ou les relations portant sur l'environnement, les utilisateurs et la notion de contexte. Ce sont évidemment des concepts que nous allons aborder dans l'ontologie que nous proposons, en adéquation avec notre cadre spécifique.

## 2.2. Ontologies pour l'assistance à l'autonomie à domicile

Comme énoncé dans notre introduction, notre projet vise à l'assistance à l'autonomie dans les déplacements de personnes à mobilité réduite. Plusieurs travaux portent sur l'assistance à l'autonomie à domicile. Malgré le contexte environnemental différent, et les différences que cela implique, il reste pertinent de nous pencher sur certains de ces travaux.

Dans ce domaine, l'ontologie présentée dans (H.K. Ngankam, 2019) porte sur les connaissances du domaine de l'informatique ubiquitaire, de la sensibilité au contexte, des capteurs et de l'environnement de la personne. Cette ontologie a été mise en place dans un système d'assistance ambiant dans les habitats intelligents pour les personnes âgées, notamment celles ayant des troubles cognitifs. Notre proposition s'appuie sur ces travaux pour représenter le volet assistance à la personne, appuyé sur la sensibilité au contexte.

Autre étude dans ce domaine, l'article (L. Chen *et al.* 2014) traite du problème de la modélisation et de la reconnaissance des activités, effectuées par les utilisateurs, au sein d'une maison intelligente (''Smart Home''). Les auteurs ont choisi une approche hybride qui profite des avantages, à la fois de l'approche fondée sur les données (fouille de données) ; et à la fois de l'approche sémantique, où une ontologie décrivant les activités permet une première heuristique. Cet article a été particulièrement inspirant pour la représentation des activités au sein de notre ontologie.

## 2.3. Ontologies pour les villes intelligentes

Plusieurs travaux, concernant des ontologies, sont orientés sur la question des villes intelligentes. Dans la plupart des cas, il s'agit de trouver une structure sémantique commune qui puisse rendre interopérables différents services au sein d'une ville.

C'est le cas pour les travaux de (P. Bellini *et al.* 2014) qui exploite une ontologie pour structurer un ensemble de données issues de plusieurs bases de données hétérogènes, dans le but de construire des services applicatifs plus variés.

De la même façon les travaux présentés dans (N. Anand *et al.* 2014) visent à structurer la logistique déployée dans une ville et est centrée sur les décideurs de la ville.

Nos travaux divergent de ces études par l'approche centrée utilisateur, décentralisée et exploitant les environnements instrumentalisés, que nous avons choisis. L'ontologie présentée dans (T. Stavropoulos *et al.* 2012) se rapproche de notre cadre applicatif et a donc été inspirant, en proposant une approche similaire, sur une échelle différente qu'est le bâtiment intelligent.

## 2.4. Ontologies pour les systèmes multi-agents

Nous avons également identifié des travaux qui font usage de systèmes structurés selon le principe multi-agents faisant usage d'ontologies.

C'est le cas du projet présenté dans (B. Gateau *et al.* 2016), qui propose de structurer sémantiquement l'environnement d'objets connectés avec lequel un système multi-agent va pouvoir interagir. La visée finale du système consiste à offrir un confort optimal par ajustement automatique du système en fonction de divers paramètres de confort, comme la température ou le niveau sonore.

D'autres travaux sur ce domaine ont cherché à structurer les échanges entre les agents d'un système multi-agents. Cela se rapproche des langages de communications entre agents (« ACL »). Dans l'article (A.R. Panisson *et al.* 2015), qui propose d'exploiter une ontologie pour gérer la négociation argumentée lors de la délégation de tâches inter-agents.

Nous avons remarqué que les ontologies développées pour les systèmes multi-agents sont souvent limitées à un contexte applicatif différent et l'approche centralisée constitue un frein à l'adaptation et la dynamicité nécessaires pour prolonger l'assistance à l'extérieur du domicile de la personne aidée dans un contexte d'informatique ambiante.

## 3. Proposition

## 3.1. Méthodologie et objectifs

Notre projet repose sur l'utilisation d'appareils connectés, qui seront exploités par un système multi-agents chargé de proposer des services d'assis-tance pour la mobilité intelligente. Nous avons évalué qu'il était nécessaire de passer par une étape de structuration sémantique des différents acteurs et objets en jeu. En effet, ces agents nécessitent donc un socle sémantique sur lequel échanger pour pouvoir offrir des services intelligents et flexibles, dans un cadre décentralisé.

La construction de cette ontologie a suivi un déroulé classique, similaire à la méthodologie décrite dans (M. Uschold et M. King, 1995). L'élaboration de l'ontologie a suivi les phases suivantes :

- **Capture de l'ontologie**, où nous définissons les termes principaux et leurs liens dans un langage naturel, à l'issue d'un ''brainstorming''.

- **Intégration avec les ontologies existantes**, où nous comparons les concepts déduits de la phase précédente avec les concepts issus d'autres travaux, notamment présentés dans la section 2. Cela permet d'en déduire les termes importants et d'en dresser une définition intermédiaire aux définitions apportées par les différents travaux.

- **Implémentation de l'ontologie**, a consisté en la transcription de l'ontologie en "OWL" avec l'outil open-source "Protégé". Nous avons choisi ce langage, puisque facilement intégrable dans notre plateforme et communs aux ontologies sur lesquelles nous nous sommes le plus basés, et donc nous faisons référence dans la section 2.

## 3.2. Concepts principaux

Les termes principaux de l'ontologie que nous présentons peuvent être observés selon 5 grands pôles, comme il est possible de le voir sur la figure 1. Nous allons aborder chaque pôle pour mieux comprendre comment ces derniers s'articulent entre eux.

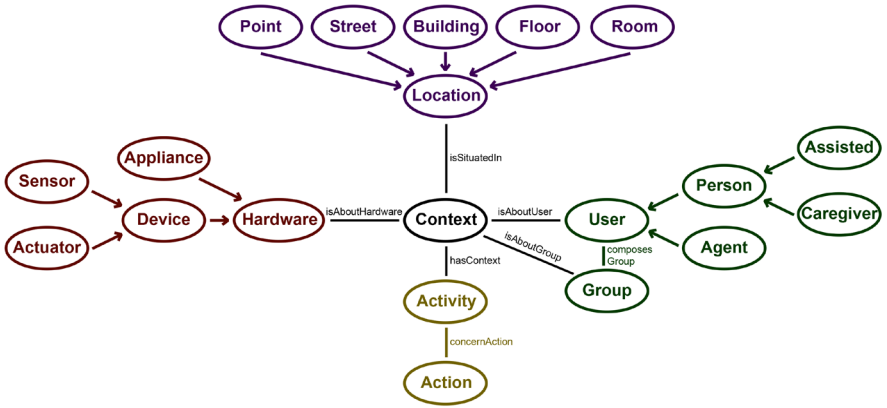

Fɪɢ. 1 – *Représentation partielle de l'ontologie,
centrée sur les concepts principaux.*

**Hardware :** La classe qui regroupe l'ensemble des appareils qui composent le système. Parmi ceux-ci, les "*Devices*" correspondent aux appareils qui peuvent influencer l'environnement physique réel. Cela regroupe à la fois les capteurs (*Sensor)*, qui peuvent collecter des informations sur des phénomènes physiques, par exemple un capteur de présence, ou un bouton. Ainsi que les effecteurs (*Actuator*), qui eux peuvent agir sur un phénomène physique, comme une lampe ou un écran. L'ensemble de ces "*Devices*" correspond finalement aux outils que les agents logiciels du système vont pouvoir utiliser pour interagir avec l'environnement physique réel et avec les utilisateurs.

D'autre part, les "*Appliance*" réunissent les autres appareils, ceux qui ne sont pas directement associés à un phénomène physique. Cela regroupe les appareils informatiques de calculs (comme un micro-ordinateur de type Raspberry-Pi), et les appareils qui fournissent une source énergétique (comme une batterie).

Il est important de préciser qu'un appareil peut être composite (*CompositeHardware*), en ce sens qu'il encapsule d'autres appareils (*Hardware*) autres que lui-même. Cela est particulièrement utile pour décrire des appareils comme les mobiles qui sont des appareils composites, comprenant plusieurs effecteurs (par exemple un écran et un haut-parleur) ainsi que plusieurs capteurs (par exemple une surface tactile et un accéléromètre).

Pour aller plus loin, cela permet aussi d'induire une logique de composition/décomposition permettant de rendre le système plus adaptatif. Par exemple, il est possible de considérer une télévision comme étant un appareil composite, composé d'un écran et d'un haut-parleur. En cas de dysfonction du téléviseur de l'utilisateur, le système, s'il en trouve dans la même pièce, pourra recomposer un téléviseur virtuel en associant un autre écran avec un haut-parleur. Cela permet également de réduire le nombre d'appareils nécessaires au fonctionnement du système, ce qui a plusieurs avantages, comme nous l'avons proposé dans (R. Fontaine *et al.*, 2020).

**User :** La classe qui contient les 2 types d'utilisateurs du système, autrement dit les entités pro-actives du système. Ce sont les entités capables d'entreprendre des *Actions.* D'une part, les personnes ("*Person*") permettent de décrire les utilisateurs humains du système. Ces derniers peuvent être des personnes aidées ("*Assisted*") ou des aidants ("*Caregiver*"). Ces derniers possèdent des caractéristiques spécifiques pour préciser la nature de l'assistance dont ils ont besoin, le type d'aide qu'ils peuvent apporter, etc. De plus, toute personne possède un profil et possiblement des préférences, qui vont permettre d'adapter le service informatique à délivrer pour permettre une cohésion et mettre en relation les personnes aidées et les aidants. D'autre part, les agents ("*Agent*") sont les autres types entités pro-actives du système, et sont logiciels. L'objectif général de ces agents est de supporter directement ou indirectement l'assistance des personnes à aider, en délivrant des services adaptés. Pour cela, les agents sont dotés d'objectifs individuels à accomplir, par exemple «alerter les secours en cas de chute de la personne». Ces objectifs seront accomplis en effectuant des *Actions*, soit qui nécessite l'utilisation d'un appareil (type "*Device*"), soit de contacter un autre utilisateur, personne ou autre agent (par exemple en demandant à cet utilisateur d'effectuer une action qu'il ne peut pas faire). Ces objectifs sont également structurés au sein de l'ontologie sous la forme d'un arbre de buts (de la même façon que le sont les appareils composites). Cette représentation des buts est en phase avec les travaux de (Braubach *et al.*, 2005 ; van Riemsdijk, Dastani et Winikoff, 2008) et donc supporte l'architecture BDI (Rao et Georgeff, 1995).

En outre, contrairement au cas des agents, le système n'a pas connaissance des objectifs individuels de la personne. Les actions sont donc déduites par le système au moyen de changements d'états de l'environnement et/ou des communications entre les agents et les personnes.

Notons que les relations sociales qu'il peut y avoir entre les utilisateurs du système peuvent également être qualifiées sur un plan qualité (satisfaisant, ambivalente, indifférente, etc.) ; et sur un plan fonctionnel (informationnelle, instrumentale et/ou émotionnelle).

La caractérisation de la relation entre les utilisateurs, et notamment la prise en compte de la nécessité de l'implication émotionnelle des acteurs dans certaines situations, permet de mieux cerner la place des Agents dans le soutien de la personne. Ces agents pourront être efficaces sur un plan instrumental, en soulageant la charge de certains aidant à cet égard. Mais ils ne peuvent pas remplacer des aidants humains, notamment lorsque cette implication émotionnelle est nécessaire.

**Location :** La classe qui regroupe les lieux permettant de situer un contexte dans un espace. Dans cette ontologie, il est possible de localiser les contextes de 2 façons : Premièrement, en indiquant des coordonnées géographiques. Cela passe par la classe "*Point*", qui permet de renseigner la latitude, la longitude et l'altitude concernée. Deuxièmement, il est possible de rattacher un lieu, avec plusieurs niveaux de granularité. À l'échelle de la ville, il est possible de préciser un quartier("*District*"), une rue ("*Street*") et un bâtiment ("*Building*"). Il est possible de descendre à plus petite échelle en précisant un étage et une salle. Ce découpage est inspiré de travaux de (X.H. Wang *et al*., 2004), (T. Stavropoulos *et al*., 2012) et (H.K. Ngankam, 2019), qui présentent une vision similaire du concept de localisation.

**Contexte :** La notion de contexte est centrale dans cette ontologie, car le contexte permet de caractériser une situation ou une activité, en répondant aux questions de qui, quoi, quand et où. Finalement, c'est l'adaptation du système au contexte qui permet d'apporter l'adaptation et la dynamicité au système. Le contexte consiste en un agrégat d'informations classées suivant leur nature (Localisation, Utilisateur(s), Objet, Temps) contextualiser un élément, souvent une activité. Pour illustrer son usage, reprenons l'exemple de la télévision présentant une dysfonction. Un agent n'aurait qu'à rechercher un écran et un haut-parleur ayant un contexte de localisation similaire (pour exemple : *Location.Building='Maison de John' ; Location.Floor=2 ; Location. Room=Bedroom*). Cette approche du concept de contexte demeure cohérente avec les propositions avancées dans (X.H. Wang *et al*., 2004) et (M. Klein *et al*., 2007) et le livre (G.D. Abowd *et al*., 1999).

**Activity :** Une activité consiste en l'occurrence d'une action (par exemple : allumer une lumière, ou faire une demande à un autre utilisateur) dans un contexte donné. Cette activité est composée d'un sujet (utilisateur qui accomplit l'action), de l'action, de l'instrument (l'outil qui permet d'accomplir l'action) et du contexte associé, comme visible sur la figure 2. Le sys-tème peut classer les activités de 2 façons : premièrement les activités prévues ("*Scheduled Activity*"), regroupe les activités qui ont été enregistrées explicitement par les sujets, ce qui concerne à la fois les activités rattachées aux personnes (par exemple si John a une sortie à la plage prévue sur son agenda), mais également les activités découlant du déroulement des objectifs des agents (par exemple si l'agent X a déclenché une alarme pour rappeler à John qu'il doit faire des courses). En outre, le système peut déduire des activités, issues de la remontée de certains motifs d'actions, on parle d'activités déduites ("*Deduced Activity*"). *A priori*, ce type d'activité ne peut avoir pour sujet que des personnes et permet, entre autres, de décrire des activités "accidentelles" issues d'incidents.

Cette représentation de l'activité dans le système est inspirée des travaux de (V. Folcher and P. Rabardel, 2004) sur les activités et le modèle de l'activité instrumentée. De plus, celle-ci reste cohérente avec les autres travaux traitants de la notion d'activité, notamment ceux de (X.H. Wang *et al*., 2004) et (L. Chen *et al*., 2014).

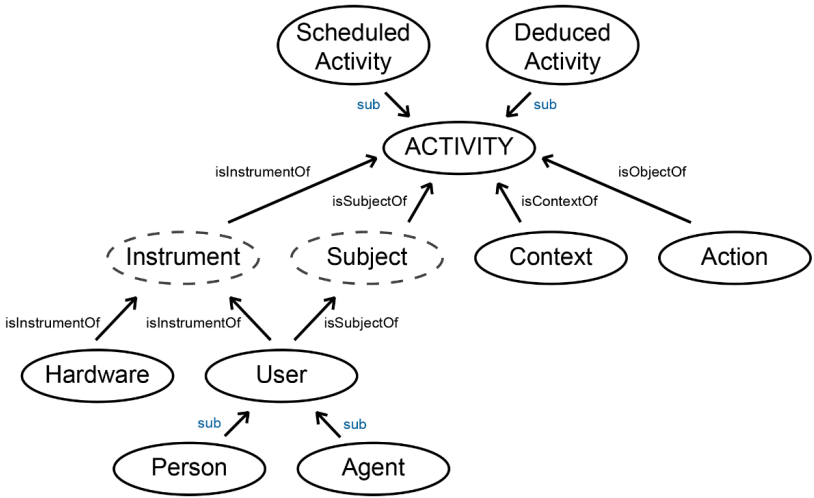

Fig. 2 – *Graphe partiel de l'ontologie, centrée sur le concept "Activity".*

### 3.3. Originalité

La première originalité de l'ontologie présentée réside dans sa transversalité, en ce sens que les concepts évoqués relèvent de plusieurs champs disciplinaires et que l'ontologie recouvre les spécificités de ces différents domaines. D'une part, l'ontologie compte une dimension assistance, donc plutôt médicale, en rapport aux utilisateurs envisagés et sur les finalités de l'usage de l'ontologie. D'autre part, il est possible de retrouver la dimension informatique dans les concepts liés au volet traitant des objets connectés et de l'informatique ambiante ainsi qu'aux agents, et donc au système multi-agents.

D'ailleurs, c'est sur ce domaine que l'ontologie propose une singularité. L'éclairage que nous proposons sur la modélisation des agents propose une vision uniforme et cohérente des entités actives du système que sont les utilisateurs humains (personnes) et les entités logicielles proactives que sont les agents. Cela passe par l'utilisation commune du concept d'activité, similaire dans les 2 groupes. Ainsi, qu'une activité (par exemple allumer une lumière) soit effectuée par un agent ou une personne, se représente de la même manière. En synergie avec la modélisation des buts des agents, cela facilite la modélisation du comportement des agents du système (Ayala *et al.*, 2019).

En outre, l'ontologie demeure légère, en étant limitée à environ 400 axiomes, pour une espace disque de 60 Ko.

## 4. Étude de cas

### 4.1. Le cas de la chute de John Doe

Nous proposons d'illustrer l'usage de cette ontologie en considérant le cas fictif de John Doe. John Doe est une personne ayant une mobilité réduite : il se déplace en fauteuil roulant.

Dans cette situation, un agent d'assistance est chargé, en cas de chute détectée, de réagir de manière adéquate. Pour cela, lorsqu'une chute est détectée, l'agent lance son comportement d'assistance. Dans un premier temps, l'utilisateur a un temps donné pour annuler l'assistance s'il estime qu'il n'en a pas besoin. Passé ce délai, ou si la personne a confirmé son besoin d'aide, l'agent va avertir un aidant proche, ainsi que prévenir les secours.

## 4.2. Implémentation dans l'ontologie

Nous allons maintenant aborder la transcription de cette situation au travers de l'ontologie présentée précédemment.

En premier lieu, l'agent requiert de détecter une chute. Cela peut se traduire en un objectif '*Détecter une chute*', qui est satisfait lorsque l'agent a la capacité d'observer une accélération à l'aide d'un 'device' ayant pour contexte 'Sensor = Accelerometer & User = John Doe'. Un mobile ou une montre connectée pourrait jouer le rôle de l'appareil requis pour mener à bien ce but.

Cet exemple est illustré, en relation avec l'ontologie, sur la figure 3 et fait écho à la notion d'adaptabilité du système, qui peut utiliser l'un ou l'autre si l'un des deux appareils présente un dysfonctionnement.

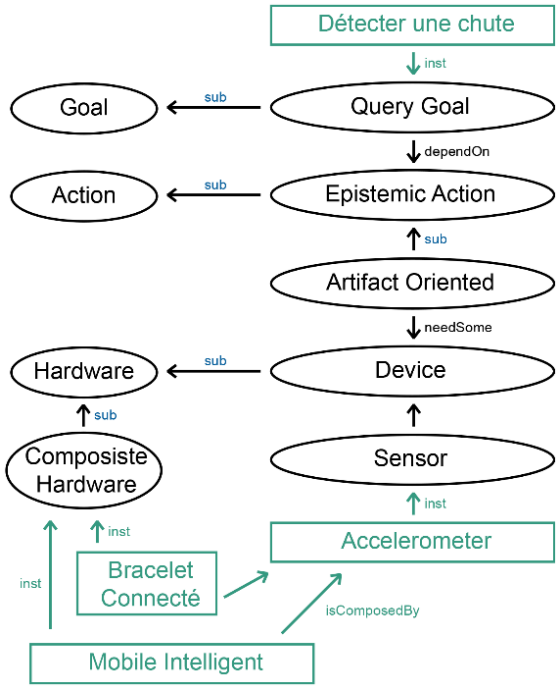

Fɪɢ. 3 – *Graphe partiel de l'action « Détecter une chute » au travers de l'ontologie, centré sur la notion de But*
*("sub" correspond à la relation "est sous-classe de...")*
*("inst" correspond à une instanciation de la classe).*

Dans un second temps, un délai est affiché à John pour annuler la séquence d'assistance. Pour cela, l'agent a pour objectif '*Proposer d'annuler l'assistance*', qui requiert l'accomplissement de 2 actions, comme visible sur la figure 3 : d'une part, l'agent doit afficher un délai pour annuler la demande, en utilisant un écran ; d'autre part, l'agent doit recevoir l'annulation de la demande, ce qui nécessite un bouton. Si le bouton n'est pas appuyé et/ou le délai n'a pas pu être affiché, on considère que la personne n'a pas annulé le processus.

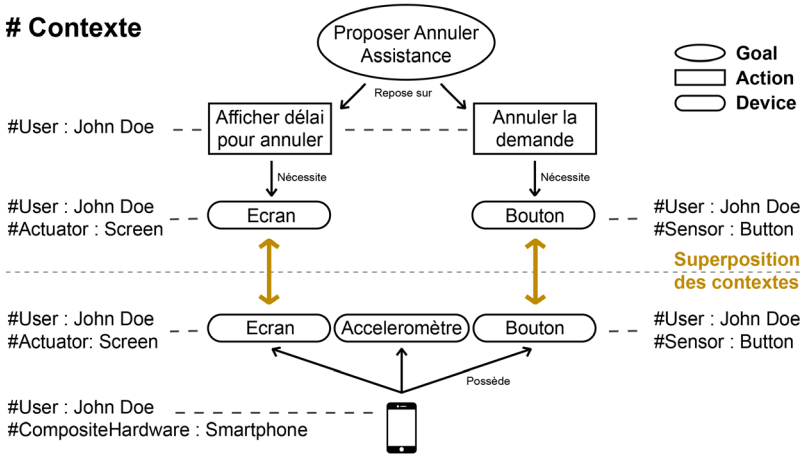

Fig. 4 – *Schéma de la superposition du contexte nécessaire pour l'accomplissement du but 'Proposer d'annuler l'assistance'.*

Celui-ci se poursuit en déclenchant la recherche d'un aidant ou, le cas échéant, l'appel aux secours professionnels.

L'accomplissement des actions nécessaires à ce but est rendu possible lors de la superposition : d'une part du contexte qui modélise les objectifs de l'agent (partie supérieure du schéma de la figure 3) ; d'autre part le contexte réel (partie inférieure), qui reflète l'état du système. En fonctionnement, lorsque l'agent devra accomplir ce but, il va rechercher un appareil au contexte validant celui qu'il recherche. Dans le cas du but '*Proposer d'annuler l'assistance*', une montre connectée, ou un mobile valide pour le contexte 'User = John Doe' délivrera les conditions nécessaires et pourront être utilisés pour effectuer les actions associées.

En d'autres termes, si la projection de l'accomplissement des objectifs est compatible avec les perceptions de l'état du système, cela a pour effet d'actionner les agents. Cette action étant située dans un contexte déterminé, elle peut être modélisée par le système comme une activité ayant eu lieu.

## 5. Conclusion et perspectives

L'ontologie présentée apporte un nouvel éclairage sur l'échange de connaissances au sein d'un système multi-agent. Ce système est destiné à l'assistance des personnes à mobilité réduite dans le cadre d'une ville intelligente. Cette dernière permet l'adaptation des comportements des agents en fonction de l'évolution de l'infrastructure et des besoins. Elle s'appuie sur d'autres ontologies reconnues, et demeure suffisamment légère pour nécessiter peu de puissance de calcul lors d'un traitement. Ce qui est une condition nécessaire dans le contexte d'un système distribué et embarqué au travers des villes.

En outre, cette ontologie permet de valider l'ensemble du fonctionnement du système en intégrant autant les connaissances du domaine informatique sur les objets connectés, et sur les agents eux-mêmes et leurs comportements, qui sont encodés au sein de l'ontologie ; ainsi que les connaissances liées à l'assistance à la personne, avec une structuration qui tourne autour de la personne aidée, et du réseau d'aidant formel et informel qui gravite autour ; et finalement sur la dynamique du système, au travers de la description des activités.

Les travaux futurs porteront sur l'intégration de l'ontologie dans notre système multi-agents, et la mise en place des raisonnements agent sur cette nouvelle base de connaissance. L'outil est ensuite destiné à être expérimenté, d'abord dans un environnement simulé, pour valider son intérêt.

Pour autant, cette ontologie demeure générique et n'est pas intriquée à notre plateforme spécifique, en ce sens qu'elle n'intègre aucune notion spécifique à notre outil, et pourrait donc être intégrée à une autre plateforme de système multi-agent ambiant.

## Remerciements

# Références

## Résumé

Nowadays, cities are equipping themselves with a large number of smart objects with a view to becoming "smart cities". To control this mass of smart objects, autonomous software entities, called agents, can be attached to them. Agents will cooperate and use these devices to offer personalized services. However, this heterogenous infrastructure needs to be semantically structured in order to be exploited. Therefore, the proposal of this paper is an ontology, formatted in OWL, describing the object infrastructures, their links with the multi-agent system organization and the services to be delivered according to the users of the system. The ontology is first applied to smart mobility for people with reduced mobility, and can be adapted to other axes of the smart city.